\documentclass{article}

\usepackage{microtype}
\usepackage{graphicx}
\usepackage{subcaption}
\usepackage{booktabs} % for professional tables

\usepackage{hyperref}

\usepackage[preprint]{icml2026}

\usepackage{amsmath}
\usepackage{amssymb}
\usepackage{mathtools}
\usepackage{amsthm}
\usepackage{xspace}
\usepackage{natbib}
\newcommand{\name}{iGVLM\xspace}
\usepackage{multirow}
\usepackage[table]{xcolor}
\usepackage{tabularray}

\usepackage[capitalize,noabbrev]{cleveref}

\theoremstyle{plain}

\theoremstyle{definition}

\theoremstyle{remark}

\usepackage[textsize=tiny]{todonotes}

\icmltitlerunning{iGVLM: Dynamic Instruction-Guided Vision Encoding for Question-Aware Multimodal Understanding}

\begin{document}

\twocolumn[
  \icmltitle{iGVLM: Dynamic Instruction-Guided \\
    Vision Encoding for Question-Aware Multimodal Understanding}



  \icmlsetsymbol{equal}{*}
  \icmlsetsymbol{intern}{\textdagger}

  \begin{icmlauthorlist}
    \icmlauthor{Hanpeng Liu}{hust,comp}
    \icmlauthor{Yaqian Li}{comp}
    \icmlauthor{Zidan Wang}{hust}
    \icmlauthor{Shuoxi Zhang}{sch}
    \icmlauthor{Zihao Bo}{comp}
    \icmlauthor{Rinyoichi Takezoe}{comp}
    \icmlauthor{Kaiwen Long}{comp}
    \icmlauthor{Kun He}{hust}
  \end{icmlauthorlist}

  \icmlaffiliation{hust}{School of Computer Science and Technology, Huazhong University of Science and Technology, Wuhan, China}
  \icmlaffiliation{comp}{Li Auto Inc.}
  \icmlaffiliation{sch}{Institute of AI for Industries, Chinese Academy of Sciences}

  \icmlcorrespondingauthor{Kaiwen Long}{longkaiwen@lixiang.com}
  \icmlcorrespondingauthor{Kun He}{brooklet60@hust.edu.cn}

  \icmlkeywords{Machine Learning, ICML}

  \vskip 0.3in
]



\printAffiliationsAndNotice{}  % no special notice (required even if empty)

\begin{abstract} Despite the success of Large Vision--Language Models (LVLMs), most existing architectures suffer from a \textbf{representation bottleneck}: they rely on \textit{static, instruction-agnostic} vision encoders whose visual representations are utilized in an invariant manner across different textual tasks. This rigidity hinders fine-grained reasoning where task-specific visual cues are critical. To address this issue, we propose \textbf{iGVLM}, a general framework for \textit{instruction-guided visual modulation}. iGVLM introduces a \textbf{decoupled dual-branch architecture}: a frozen representation branch that preserves task-agnostic visual representations learned during pre-training, and a dynamic conditioning branch that performs \textbf{affine feature modulation} via Adaptive Layer Normalization (AdaLN). This design enables a smooth transition from general-purpose perception to instruction-aware reasoning while maintaining the structural integrity and stability of pre-trained visual priors. Beyond standard benchmarks, we introduce \textbf{MM4}, a \textbf{controlled diagnostic probe} for quantifying \textit{logical consistency} under multi-query, multi-instruction settings. Extensive results show that iGVLM consistently enhances instruction sensitivity across diverse language backbones, offering a plug-and-play paradigm for bridging passive perception and active reasoning. \end{abstract}
\section{Introduction}
\begin{figure}[t!]
\vskip 0.2in
  % \vspace{-0.1em}
\centering
\includegraphics[width=0.8\linewidth]{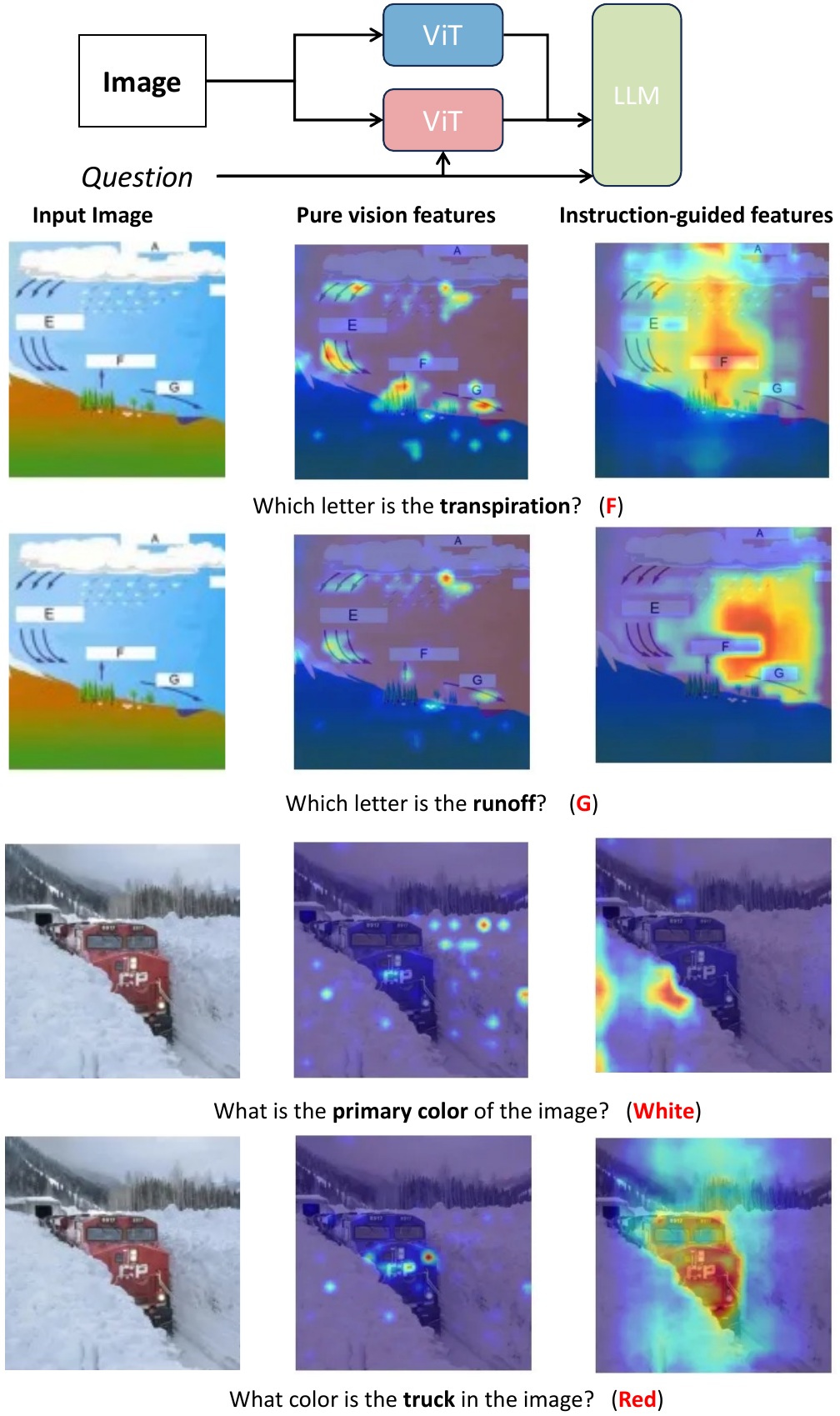}      
% \caption{\textbf{
% %Architectural differences in VLMs.
% Architectural Comparison of Vision-Language Models.} 
% Top: Pipeline of LLaVA-1.5.
% Middle: Pipeline of QA-ViT.
% Bottom: Pipeline of our proposed \textit{\name}. 
% %We employ a dual-branch to fuse pure vision features with instruction-guided features.
% Unlike prior approaches, \textit{\name} employs a dual-branch vision encoder that fuses static visual features with dynamically modulated, instruction-guided features for enhanced instruction-aware perception.
% }
\caption{\textbf{Visualization of vision features.} We employ Grad-CAM to visualize the vision encoders of the two branches in iGVLM, highlighting the regions most relevant to the correct answer. As shown in the figure, the instruction-guided branch distinctly focuses on areas that are more closely associated with the correct answer.
}
\label{fig:visualization}
% \vspace{-2em}
\end{figure}

In recent years, advances in computer vision~\cite{DBLP:conf/cvpr/ZhangL0024,DBLP:conf/iccv/LiuL00W0LG21} and natural language processing~\cite{DBLP:conf/nips/VaswaniSPUJGKP17,radford2019language,brown2020language} have driven remarkable progress in Vision--Language Models (VLMs)~\cite{sharegpt4v,DBLP:journals/corr/abs-2403-05525,DBLP:journals/corr/abs-2312-14238,DBLP:journals/corr/abs-2401-04088}. 
By jointly modeling visual perception and linguistic understanding, these models achieve strong performance on multimodal tasks such as image captioning, visual question answering, and grounded dialogue, representing an important step toward general-purpose multimodal intelligence. 
Despite this progress, a fundamental challenge remains: how to condition visual perception on task-specific linguistic instructions in a principled and efficient manner.

Most existing VLMs rely on \emph{static, instruction-agnostic} vision encoders, such as CLIP-ViT~\cite{DBLP:conf/icml/RadfordKHRGASAM21}, which extract visual representations independently of the downstream textual query.
As a result, visual features are reused across different instructions in an invariant manner, limiting the model’s ability to emphasize task-relevant cues and perform fine-grained, question-aware reasoning.
This limitation is qualitatively illustrated in~\cref{fig:visualization}, where static visual representations fail to highlight instruction-dependent regions that are critical for answering different questions grounded in the same image.
These observations suggest that the core difficulty lies not in relearning visual perception itself, but in conditioning the \emph{utilization} of visual features on linguistic instructions.

%##################################################################################################
\begin{figure*}[t]\centering
\includegraphics[width=0.9\linewidth]{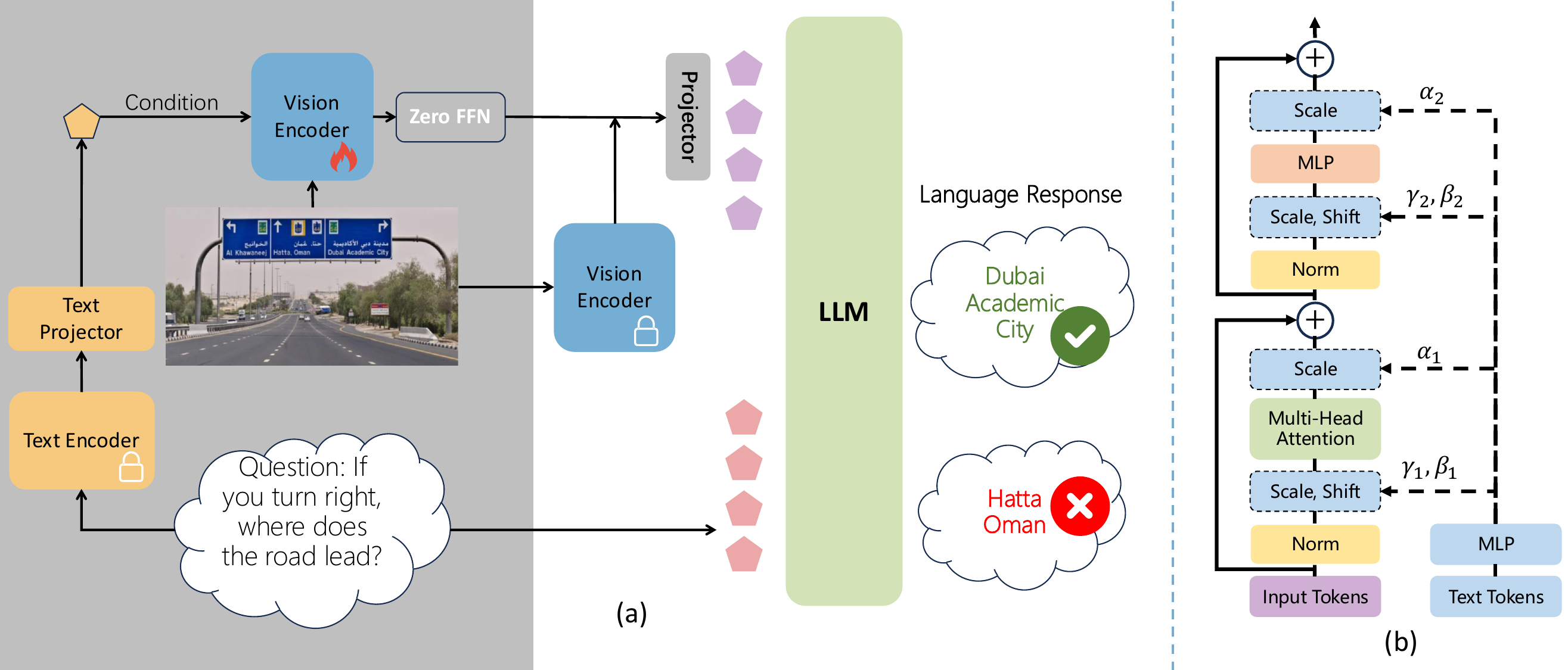}
    \caption{\textbf{(a): The proposed \name architecture.} The Text Encoder extracts features from the input instructions to guide the Vision Encoder, %dynamically adjusting the visual features. 
    enabling dynamic modulation of visual representations.
    These instruction-conditioned features are then fused with static visual features. 
    The fused representation is aligned with a Large Language Model (LLM) to generate responses. 
    The illustrated example comes from a real-world VQA scenario rather than the MM4 benchmark. 
    \textbf{(b): AdaLN-Modified ViT.} We leverage textual information to modulate the multi-head attention and MLP modules within the ViT the AdaLN adapter, %thereby directing the evolution of visual focus.
    enabling instruction-aware adjustment of visual attention.
}
\label{fig:mechanisms}
\vspace{-1em}
\end{figure*}
%##################################################################################################

Consequently, recent work has explored lightweight mechanisms to introduce instruction awareness while preserving the perceptual strength of pretrained vision encoders.
QA-ViT~\cite{DBLP:conf/cvpr/GanzKABNML24} injects textual representations into upper layers of a frozen vision transformer, enabling limited instruction-dependent adaptation with high efficiency.
However, such partial integration provides relatively weak conditioning and may still perturb pretrained visual representations.
In contrast, DyFo~\cite{dyfo} formulates visual reasoning as a sequential decision process guided by external expert models and Monte Carlo Tree Search, allowing more flexible, instruction-aware attention shifts at the cost of substantial inference overhead and reliance on expert quality.
Taken together, existing approaches highlight the challenge of achieving effective instruction conditioning while maintaining both computational efficiency and representation stability.

Motivated by this observation, we propose \textbf{iGVLM}, a \emph{decoupled instruction-guided vision encoder} for Vision--Language Models.
iGVLM adopts a dual-branch architecture that separates static and dynamic perception pathways: a frozen static branch preserves task-agnostic visual representations learned during pre-training, while a dynamic branch integrates lightweight, instruction-conditioned adapter modules that modulate feature utilization under textual guidance.
This design enables flexible, instruction-aware visual reasoning without retraining the backbone, achieving a favorable balance between adaptability, efficiency, and representation stability.
We evaluate iGVLM on the MMStar~\cite{DBLP:conf/nips/ChenLDZZCDWQLZ24} benchmark for fine-grained multimodal reasoning, and further introduce \textbf{MM4}, a controlled diagnostic benchmark for assessing question-aware visual reasoning under multi-query, multi-instruction settings.

Our main contributions are summarized as follows: \begin{itemize} 
\item We propose \textbf{iGVLM}, a decoupled instruction-guided vision encoder that separates representation preservation from instruction-conditioned adaptation via a dual-branch architecture. 

\item We introduce \textbf{MM4}, a controlled diagnostic benchmark for evaluating question-aware visual perception under multi-query, multi-instruction scenarios. 

\item We demonstrate through extensive experiments on MMStar and other multimodal benchmarks that iGVLM improves instruction sensitivity and fine-grained reasoning while maintaining efficiency and general-purpose multimodal capability. 

\end{itemize}
\section{Related work}
\label{sec:relate}

\paragraph{Vision Encoders in Vision--Language Models.}
Recent years have witnessed rapid progress in Vision--Language Models (VLMs)~\cite{DBLP:journals/corr/abs-2501-17811,DBLP:journals/corr/abs-2402-14289,DBLP:journals/corr/abs-2412-10302}, driven by advances in both large-scale multimodal pretraining and architectural design.
A foundational line of work, exemplified by CLIP~\cite{DBLP:conf/icml/RadfordKHRGASAM21}, demonstrates that contrastive learning on large-scale image--text pairs can effectively align visual and textual representations, forming the basis of many modern VLMs.
Subsequent studies have explored how to enhance the visual encoding component within VLMs to better support downstream multimodal reasoning.

One line of research focuses on strengthening visual representations by aggregating information from multiple encoders or pretrained visual models.
For example, Ranzinger et al.~\cite{ranzinger2024radio} fuse features from multiple vision encoders, while Tong et al.~\cite{tong2024eyes} augment CLIP features with representations from DINOv2~\cite{oquab2023dinov2}, leading to improved visual grounding.
Monkey~\cite{li2024monkey} further explores fine-tuning multiple vision encoders to support high-resolution image understanding.
These approaches primarily aim to improve the \emph{capacity} and \emph{coverage} of visual representations, but typically rely on static encoders whose outputs are invariant to task-specific instructions. A complementary line of work investigates how to introduce query- or instruction-awareness into the vision encoder.
QA-ViT~\cite{DBLP:conf/cvpr/GanzKABNML24} incorporates query-aware cross-attention to modulate visual features based on textual prompts, enabling more effective integration of visual and linguistic information for question answering.
While such designs provide a degree of instruction-dependent adaptation, they often operate within a single encoder pathway and offer limited control over how pretrained visual representations are preserved or modified.
In contrast to these approaches, our work focuses on explicitly decoupling representation preservation from instruction-conditioned modulation within the vision encoder.

\paragraph{Evaluating Vision--Language Models.}
Evaluating the capabilities of VLMs has been an active area of research, leading to the development of a diverse set of multimodal benchmarks.
Early benchmarks, such as VQA~\cite{DBLP:conf/cvpr/GoyalKSBP17}, MS-COCO~\cite{DBLP:conf/acl/SoricutDSG18}, and OK-VQA~\cite{DBLP:conf/eccv/SchwenkKCMM22}, provide task-specific assessments of multimodal perception and reasoning.
More recent efforts aim to offer broader and more challenging evaluations of multimodal understanding and instruction following~\cite{DBLP:conf/iclr/0001Z0CLWLSYZL24,DBLP:journals/corr/abs-2306-13394,DBLP:journals/corr/abs-2311-15596}.
MMStar~\cite{DBLP:conf/nips/ChenLDZZCDWQLZ24} further consolidates existing benchmarks and introduces a carefully curated, vision-dependent evaluation suite designed to mitigate data leakage and spurious correlations.

Despite these advances, most existing benchmarks primarily assess \emph{general-purpose} multimodal capabilities and evaluate each query in isolation.
As a result, they provide limited insight into whether a model can consistently adapt its visual perception to different instructions grounded in the same image.
To address this gap, we introduce \textbf{MM4}, a controlled diagnostic benchmark specifically designed to evaluate \emph{question-aware} visual understanding.
MM4 challenges models to answer multiple, semantically distinct queries associated with a single image, enabling more fine-grained analysis of instruction-conditioned visual perception and multi-query consistency.

\section{Method}

In this section, we introduce \textbf{iGVLM}, a decoupled instruction-guided vision encoder designed to condition the utilization of visual features on linguistic instructions while preserving pretrained visual representations.
We first present an overview of the overall architecture, followed by detailed descriptions of (i) instruction-guided visual feature modulation and (ii) dual-branch feature fusion.
Finally, we introduce MM4, a controlled diagnostic benchmark for evaluating question-aware visual perception in Vision--Language Models (VLMs).

\subsection{Overall Architecture}

An overview of the proposed framework is illustrated in~\cref{fig:mechanisms}(a).
Given an image--text pair, iGVLM conditions visual feature generation on the textual instruction through a dedicated conditioning pathway, while preserving the original perceptual capacity of the pretrained vision backbone.
Specifically, the textual instruction is first encoded into a compact semantic representation, which serves as a global guidance signal for visual modulation.
This instruction embedding conditions a pretrained vision encoder, enabling visual features to be selectively modulated according to task-specific linguistic cues.

To explicitly separate representation preservation from instruction-conditioned adaptation, iGVLM adopts a dual-branch architecture.
A static branch retains a frozen vision encoder to preserve task-agnostic visual priors, while a dynamic branch generates instruction-adapted visual features through lightweight modulation modules.
The outputs of these two branches are fused to obtain a balanced visual representation that combines general-purpose perceptual semantics with task-specific adaptation.
The fused visual features are subsequently projected into the language embedding space and provided, together with the instruction tokens, to a large language model (LLM) for multimodal reasoning and response generation.

\subsection{Instruction-Guided Visual Feature Modulation}

To enable instruction-conditioned visual perception, we derive a global textual guidance signal from the instruction.
We adopt the text encoder from a pretrained CLIP model and truncate the input text to a maximum length of 77 tokens.
The resulting [CLS] token embedding summarizes the semantic intent of the instruction and is mapped into the vision latent space through a lightweight linear projection:
\begin{equation}
    c_t = \mathcal{F}_T(T_{\leq 77}), \quad
    \hat{c}_t = \mathcal{H}_t(\mathrm{Norm}(c_t)),
\end{equation}
where $\mathcal{F}_T(\cdot)$ denotes the CLIP text encoder, and $\mathcal{H}_t(\cdot)$ aligns the text embedding with the vision feature space.

We incorporate Adaptive Layer Normalization (AdaLN)~\cite{DBLP:conf/aaai/PerezSVDC18} into each transformer block of the CLIP vision encoder to inject textual conditioning in a stable and localized manner.
The projected instruction embedding $\hat{c}_t$ is transformed into layer-wise modulation parameters that control feature scaling and shifting within both the self-attention and feedforward submodules.
By integrating AdaLN across all transformer layers, iGVLM enables hierarchical instruction-conditioned modulation while preserving the pretrained weights of the vision backbone.

Formally, given an input image $I$ and instruction embedding $\hat{c}_t$, the instruction-guided vision encoder produces:
\begin{equation}
    y_{ct} = \mathcal{F}_{ct}(I, \hat{c}_t; \Theta_{\mathrm{CLIP}}),
\end{equation}
where $y_{ct} \in \mathbb{R}^{N_I \times D_I}$ denotes the instruction-conditioned visual features and $\Theta_{\mathrm{CLIP}}$ represents the frozen pretrained parameters.

\subsection{Dual-Branch Feature Fusion}

While instruction-conditioned modulation enables task-specific adaptation, preserving the original perceptual semantics is essential for stable and generalizable visual understanding.
To this end, iGVLM employs a dual-branch fusion mechanism that explicitly combines instruction-guided features with the original frozen visual representations.

Let $y_{ct} \in \mathbb{R}^{N_I \times D_I}$ denote the instruction-guided features from $\mathcal{F}_{ct}$, and let $y_0 = \mathcal{F}_I(I; \Theta_{\mathrm{CLIP}})$ denote the corresponding frozen features from the original vision encoder.
The fused visual representation is computed as:
\begin{equation}
    y_I = \mathcal{Z}(\mathrm{Norm}(y_{ct})) + y_0,
\end{equation}
where $\mathcal{Z}$ is a learnable linear projection initialized to zero.
This initialization ensures that the fused representation initially matches the pretrained visual features, allowing instruction-conditioned adaptation to be introduced gradually and safely during training.

Following the LLaVA-1.5 framework, the fused visual features $y_I$ are projected into the input embedding space of the LLM via a learnable linear transformation.
Training proceeds in two stages: first, the instruction-guided vision encoder and projection layers are optimized while keeping the pretrained vision backbone and LLM frozen; second, all components are jointly optimized to enable coherent multimodal reasoning.

\subsection{MM4: A Diagnostic Benchmark for Question-Aware Visual Perception}

To complement existing multimodal benchmarks such as MMStar~\cite{DBLP:conf/nips/ChenLDZZCDWQLZ24}, which primarily assess general-purpose multimodal reasoning, we introduce \textbf{MM4}, a controlled diagnostic benchmark designed to evaluate \emph{question-aware} and \emph{multi-query} visual perception.
MM4 consists of 180 images and 720 manually verified question--answer pairs, with annotations curated by domain experts to ensure quality and consistency.

Each image in MM4 is associated with four semantically distinct questions, constructed according to three design principles:
(i) robustness through answer reversal, (ii) multi-perspective semantic diversity, and (iii) balanced answer distribution.
This design enables MM4 to jointly assess intra-image consistency and inter-question diversity.
To evaluate multi-query reasoning, MM4 adopts a hierarchical scoring protocol that credits a model only if it correctly answers at least $n$ of the four questions per image, encouraging consistent instruction-aware reasoning rather than isolated accuracy.

\section{Experiments}
\subsection{Experimental Settings}

\paragraph{Training Setup.}
All experiments are conducted within the LLaVA-1.5 training framework to ensure a controlled and fair comparison. We use the same open-source training data as LLaVA-1.5, including 558K image--text pairs for alignment pretraining and 665K samples for instruction tuning, without introducing any additional data. All models share the same visual backbone, \textbf{CLIP-Large-336}~\cite{DBLP:conf/icml/RadfordKHRGASAM21}, and differ only in how visual features are conditioned on textual instructions. We evaluate \name\ across multiple language backbones, including \textbf{Vicuna-7B}, \textbf{Vicuna-13B}~\cite{vicuna}, and \textbf{Qwen2.5-3B}~\cite{qwen2.5}, enabling analysis of both intra-family scaling and cross-architecture generalization. All experiments are run on 8 NVIDIA A100 GPUs under identical hardware and software configurations. Compared to LLaVA-1.5, training a 7B version of \name\ incurs a moderate computational overhead of approximately \textbf{1.1$\times$} GPU hours, reflecting the lightweight nature of the proposed instruction-guided visual modulation.

\paragraph{Evaluation Benchmarks.}
Our primary evaluation is conducted on MMStar~\cite{DBLP:conf/nips/ChenLDZZCDWQLZ24}, a vision-dependent multimodal benchmark designed to assess fine-grained reasoning while minimizing data leakage, and we report MMStar results as the main indicator of general-purpose multimodal performance. To further examine instruction sensitivity and generalization, we additionally evaluate on a range of established benchmarks, including VQAv2~\cite{DBLP:conf/cvpr/GoyalKSBP17}, GQA~\cite{DBLP:conf/cvpr/HudsonM19}, POPE~\cite{DBLP:conf/emnlp/LiDZWZW23}, VizWiz~\cite{DBLP:conf/cvpr/Gurari0SGLGLB18}, and ScienceQA~\cite{DBLP:conf/nips/LuMX0CZTCK22}, which collectively assess open-ended visual understanding, robustness to hallucination, zero-shot generalization, and scientific reasoning.

\paragraph{Baselines.}
We adopt \textbf{LLaVA-1.5}~\cite{llava-1.5} as the primary baseline, as it provides fully open-source data, model weights, and training code. All baseline models are trained using the same data, optimization settings, and vision backbone as \name\, ensuring that performance differences arise solely from differences in visual encoding strategies. For Vicuna-based backbones, we additionally compare with representative instruction-aware modulation methods, including \textbf{QA-ViT}~\cite{DBLP:conf/cvpr/GanzKABNML24} and \textbf{DyFo}~\cite{dyfo}, which introduce query-aware modulation and expert-guided search, respectively. For Qwen2.5-3B, we compare against LLaVA-1.5 with the same backbone due to architectural compatibility, ensuring consistent evaluation across different model families and instruction-conditioning strategies.

\begin{table*}[t!]
\centering
% \caption{\textbf{Comparisons with other dynamic visual perception methods on MMStar .} Our \name achieves the best overall performance.}
\caption{
Comparison among dynamic visual perception methods on \textbf{MMStar}, 
evaluated across six perception dimensions (CP, FP, IR, LR, ST, MA), 
average accuracy (Avg.), and throughput (it/s). 
Our \textbf{iGVLM} consistently outperforms prior approaches under both 
\textbf{Vicuna} and \textbf{Qwen2.5} backbones, achieving the best overall 
balance between accuracy and efficiency. $\Delta$ indicates improvement over LLaVA-1.5 under the same backbone. \textbf{Bold} and \underline{underlined} values indicate the best and second-best results, respectively.
}
\label{tab:main}
% \vspace{-1em}
\scalebox{0.9}{
\begin{tabular}{l l   | c  c c c  c  c  |c c | c}
\toprule
Method& Venue  & CP & FP & IR & LR & ST & MA & Avg. &  ${\Delta}$& Thpt \\
\midrule
\multicolumn{11}{c}{\cellcolor{gray!10}\textbf{Vicuna-7B Backbone}} \\
LLaVA-1.5 & CVPR'24  & 58.8 & 24.0 & 38.8 & 24.0 & 13.6 & 22.8 & 30.3 & --& 13.5\\
QA-ViT & CVPR'24 & \textbf{60.0} & 28.0 & 40.4 & 25.6 & 14.0 & 21.2 & 31.5& $+1.2$ & 12.9\\
DyFo & CVPR'25  & 54.8 & \textbf{28.4} & 38.0 & 25.6 & \textbf{24.0} & \textbf{27.2} & 33.0 & $+2.7$ & 0.49\\
\textbf{\name} & Ours  & \underline{59.6} & \underline{28.0} & \textbf{41.2} & \textbf{32.8} & \underline{19.6} & \underline{26.8} & \textbf{34.7} &   \textbf{+4.4}&  11.1 \\
\midrule
\multicolumn{11}{c}{\cellcolor{gray!10}\textbf{Vicuna-13B Backbone}} \\
LLaVA-1.5 & CVPR'24 & \textbf{58.8} & 28.0 & 41.6 & 24.4 & 18.4 & 25.6 & 32.8 & --  &  9.8 \\
QA-ViT & CVPR'24  & 57.6 & 32.8 & 43.2 & \textbf{28.8} & 21.2 & \textbf{31.6} & 35.9 & $+3.1$  & 10.2 \\
DyFo & CVPR'25   & 55.6 & 32.0 & \textbf{44.4} & 27.2 & 22.0 & 26.4 & 34.6 & $+1.8$  & 0.47\\
\textbf{\name} & Ours&  \underline{57.6} & \textbf{37.6} & \underline{43.2} & \underline{27.2} & \textbf{22.4} & \underline{30.8} & \textbf{36.4} & \textbf{+3.6} & 8.6 \\
% \textbf{\name-dul} & & Vicuna-13B & 56.0 & 32.4 & \underline{43.6} & \textbf{28.8} & 19.6 & 30.4 & 35.1 & 2:54 & 8.6\\
\midrule
\multicolumn{11}{c}{\cellcolor{gray!10}\textbf{Qwen2.5-3B Backbone}} \\
LLaVA-1.5 & CVPR'24 & 8.4&	8.4	&16.4	&24.4	&16.8	&26.4&	16.8 & --&  12.6\\
\textbf{\name} & Ours & \textbf{20.0}	& \textbf{13.2}	&\textbf{19.2}  & \textbf{28.4}	&\textbf{21.6}	& 25.6 & \textbf{21.3} & \textbf{+4.5}  &  10.9\\
\bottomrule
\end{tabular}
}
% \vspace{-1em}
% \vskip -0.1in
\end{table*}
\subsection{Results on MMStar}

We evaluate iGVLM on MMStar under three representative settings.
For Vicuna-based models (7B and 13B), we compare iGVLM with both the static baseline LLaVA-1.5 and two instruction-aware modulation methods, QA-ViT and DyFo.
For the Qwen2.5-3B backbone, we compare against the corresponding LLaVA-1.5 model due to architectural compatibility.
All methods are evaluated using identical training data, vision backbones, and inference configurations to ensure a controlled comparison.

As shown in~\cref{tab:main}, iGVLM achieves the best overall performance across all backbones.
On Vicuna-7B, iGVLM improves the average MMStar score by \textbf{+4.4} points over LLaVA-1.5, outperforming both QA-ViT (+1.2) and DyFo (+2.7).
Notably, the gains are most pronounced on instruction-sensitive and fine-grained reasoning dimensions, including Instance Reasoning (IR), Logical Reasoning (LR), and Science \& Technology (ST), where iGVLM consistently surpasses both baselines.
A similar trend is observed on Vicuna-13B: iGVLM achieves an average improvement of \textbf{+3.6} points, exceeding QA-ViT (+3.1) and DyFo (+1.8), with particularly strong gains on Fine-grained Perception (FP) and ST.
These results indicate that explicitly conditioning the utilization of visual features enables more effective instruction-aware reasoning than single-path modulation (QA-ViT) or expert-guided search (DyFo).

In addition to accuracy, iGVLM maintains a favorable efficiency profile.
Despite achieving higher overall performance, iGVLM preserves throughput comparable to LLaVA-1.5, with only a modest reduction (13.5$\rightarrow$11.1 for Vicuna-7B and 9.8$\rightarrow$8.6 for Vicuna-13B).
In contrast, DyFo incurs a severe efficiency penalty due to repeated expert-guided search, reducing throughput by more than \textbf{20$\times$} (13.5$\rightarrow$0.49 for Vicuna-7B and 9.8$\rightarrow$0.47 for Vicuna-13B).
QA-ViT maintains efficiency similar to LLaVA-1.5, but achieves only limited accuracy gains, highlighting the trade-off between conditioning strength and computational cost in existing approaches.

Results on Qwen2.5-3B further confirm the generality of iGVLM.
Compared to LLaVA-1.5, iGVLM improves the average MMStar score from 16.8 to \textbf{21.3} (+4.5), with consistent gains across all capability dimensions.
Taken together, these results demonstrate that iGVLM strikes a more effective balance between instruction-aware reasoning and computational efficiency than prior dynamic modulation methods, validating the advantages of decoupled instruction-guided visual encoding on a general-purpose multimodal benchmark.

% \begin{figure}[t!]
% \vskip 0.2in
%   \centering
%  \includegraphics[width=1\linewidth]{figure/MM4_Model_Scores.pdf} 
%  \caption{\textbf{MM4 Benchmark Leaderboard.} GPT-4o achieves state-of-the-art performance, while our iGVLM-13B attains competitive accuracy comparable to GPT-4v.}
% \label{fig:mm4}
% % \vspace{-1em}
% \end{figure}

\subsection{Results on MM4}

We evaluate iGVLM on the proposed MM4 benchmark, which is specifically designed to assess \emph{question-aware} and \emph{multi-query} visual reasoning under shared visual inputs.
Unlike general-purpose benchmarks that evaluate each query in isolation, MM4 requires models to adapt visual perception consistently across multiple, semantically distinct questions grounded in the same image.

\textbf{Quantitative Results.}
As shown in~\Cref{tab::mm4_s}, closed-source systems such as GPT-4o~\cite{gpt-4o} and Qwen2.5-vl-max~\cite{qwen2.5vl} achieve the highest absolute scores, reflecting the advantages of large-scale proprietary models.
Among open-source systems, iGVLM consistently outperforms its corresponding LLaVA-1.5 baselines under the same backbone.
In particular, \textbf{iGVLM-3B} achieves the best performance among open-source models, improving over LLaVA-1.5-3B despite having the same parameter scale.
Notably, iGVLM-3B also outperforms the larger iGVLM-13B, indicating that MM4 performance is driven more by instruction-aware visual utilization than by parameter count alone.
These results suggest that the proposed decoupled visual modulation can effectively leverage stronger language backbones without architectural modification.

\textbf{Multi-Query Consistency Analysis.}
MM4 adopts an increasingly strict evaluation protocol in which a model receives credit only if it correctly answers at least $n$ out of four questions per image ($n=1,2,3,4$).
As reported in~\Cref{tab::mm4_s}, performance decreases monotonically as $n$ increases for all models, reflecting the growing difficulty of maintaining consistent reasoning across multiple queries.
However, iGVLM exhibits a noticeably slower performance degradation compared to baseline methods, particularly at higher consistency thresholds ($n=3$ and $n=4$).
This behavior indicates that instruction-guided visual modulation enables more stable adaptation of visual attention across different questions, rather than relying on isolated correct predictions.

To contextualize these results, we note that random guessing yields an expected score of approximately $0.7$ at $n=4$, derived from a per-question accuracy of $0.25$ over four independent questions.
All evaluated models perform substantially above this baseline, confirming that MM4 provides a discriminative evaluation of multi-query reasoning.
Importantly, the relative advantage of iGVLM becomes more pronounced under stricter consistency requirements, aligning with the intended diagnostic goal of MM4.

\begin{figure*}[t!]
\vskip 0.2in
\centering
\includegraphics[width=0.8\linewidth]{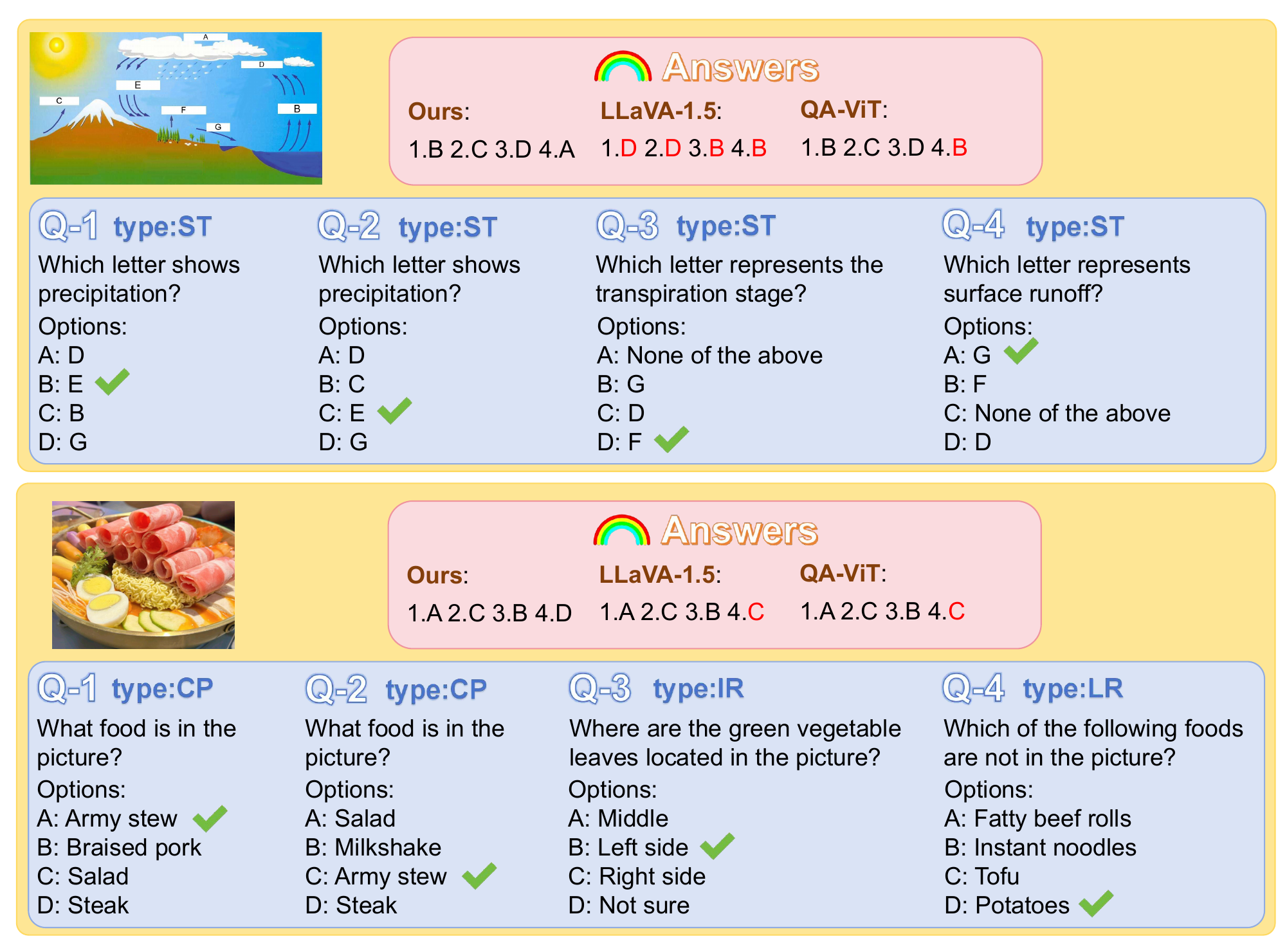}
%\vspace{.1em}
% \vspace{-1em}
\caption{\textbf{Representative Examples from MM4.} Deceptively simple questions demand multiperspective reasoning. Correct answers (highlighted) are uniformly distributed across options to prevent positional bias. More examples can be found in~\cref{sec:cases}.
}
% \vspace{-1em}
\label{fig:exp}
\end{figure*}

\textbf{Qualitative Analysis.}
We further visualize representative examples in~\cref{fig:exp} to illustrate how instruction-guided visual modulation affects model behavior.
Compared with LLaVA-1.5-13B and QA-ViT-13B, iGVLM-13B more accurately localizes instruction-relevant regions under different queries.
In the science diagram example, iGVLM distinguishes semantically similar stages such as \emph{evaporation} and \emph{transpiration}, while baseline models attend to ambiguous regions.
In the food scene example, iGVLM demonstrates stronger compositional reasoning by correctly identifying missing objects and spatial relationships across multiple questions.
Together, these qualitative and quantitative results confirm that iGVLM enhances question-aware visual perception by enabling consistent, instruction-conditioned feature utilization.

\subsection{Other Benchmarks}

To assess whether instruction-guided visual modulation affects general-purpose multimodal capabilities, we further evaluate \textbf{iGVLM} on a diverse set of established benchmarks, including VQAv2, GQA, POPE, VizWiz, and ScienceQA-IMG, as summarized in~\cref{tab:qa}. 
These benchmarks cover complementary aspects of vision--language understanding, ranging from open-ended visual reasoning (VQAv2, GQA), hallucination robustness (POPE), real-world visual grounding (VizWiz), to domain-specific scientific reasoning (ScienceQA-IMG). In contrast, \textbf{DyFo} relies heavily on Monte Carlo Tree Search (MCTS) and lacks specialized search strategies for non-selective benchmarks such as VQAv2, GQA, and VizWiz, which significantly limits its generalizability compared to QA-ViT and our proposed \textbf{iGVLM}.

As shown in~\cref{tab:qa}, \textbf{iGVLM} consistently maintains comparable or improved performance relative to \textbf{LLaVA-1.5} across different model scales and backbones.
For Vicuna-based models, iGVLM yields modest but consistent gains on most benchmarks.
In particular, iGVLM improves POPE accuracy from 85.4 to 85.9 on Vicuna-7B and from 85.4 to 86.1 on Vicuna-13B, indicating enhanced robustness against visual hallucination.
Notable improvements are also observed on VizWiz, where iGVLM raises accuracy from 50.0 to 52.5 for Vicuna-7B and from 53.6 to 55.3 for Vicuna-13B, suggesting more reliable visual grounding under real-world conditions.
On ScienceQA-IMG, iGVLM achieves clear gains for Vicuna-7B (+3.1), while maintaining comparable performance for Vicuna-13B.

Under the Qwen2.5-3B backbone, iGVLM shows a similar trend.
While performance on VQAv2 and GQA remains comparable to LLaVA-1.5-3B, iGVLM substantially improves VizWiz accuracy from 50.7 to 53.4 and ScienceQA-IMG accuracy from 72.2 to 73.0.
Across all evaluated benchmarks, no systematic performance degradation is observed, indicating that instruction-guided visual modulation does not compromise general-purpose multimodal reasoning.
Overall, these results suggest that iGVLM serves as a drop-in enhancement to existing vision--language models, improving instruction-aware visual utilization while preserving broad applicability across diverse multimodal tasks.

\begin{table*}[t!]
\centering
\caption{\textbf{Scores of different methods on MM4.} Different values of $n$ indicate different scoring schemes.
}
\label{tab::mm4_s}
% \vspace{-1em}
\resizebox{\textwidth}{!}{
\begin{tblr}{
  cell{2}{2} = {c},
  cell{2}{3} = {c},
  cell{2}{4} = {c},
  cell{2}{5} = {c},
  cell{2}{6} = {c},
  cell{2}{7} = {c},
  cell{2}{8} = {c},
  cell{2}{9} = {c},
  cell{2}{10} = {c},
  cell{2}{11} = {c},
  cell{2}{12} = {c},
  cell{3}{2} = {c},
  cell{3}{3} = {c},
  cell{3}{4} = {c},
  cell{3}{5} = {c},
  cell{3}{6} = {c},
  cell{3}{7} = {c},
  cell{3}{8} = {c},
  cell{3}{9} = {c},
  cell{3}{10} = {c},
  cell{3}{11} = {c},
  cell{3}{12} = {c},
  cell{4}{2} = {c},
  cell{4}{3} = {c},
  cell{4}{4} = {c},
  cell{4}{5} = {c},
  cell{4}{6} = {c},
  cell{4}{7} = {c},
  cell{4}{8} = {c},
  cell{4}{9} = {c},
  cell{4}{10} = {c},
  cell{4}{11} = {c},
  cell{4}{12} = {c},
  cell{5}{2} = {c},
  cell{5}{3} = {c},
  cell{5}{4} = {c},
  cell{5}{5} = {c},
  cell{5}{6} = {c},
  cell{5}{7} = {c},
  cell{5}{8} = {c},
  cell{5}{9} = {c},
  cell{5}{10} = {c},
  cell{5}{11} = {c},
  cell{5}{12} = {c},
  hline{2} = {-}{},
  hline{1, 6}={-}{1pt},
}
$n$ & GPT-4o & Qwen2.5-vl-max & GPT-4v & iGVLM-13B & QA-ViT-13B & LLaVA1.5-13B & iGVLM-7B & LLaVA1.5-7B & QA-ViT-7B & LLaVA-1.5-3B & iGVLM-3B \\
1    & 170     & \textbf{172}           & 161     & 161       & 163        & 161          & 161      & 157         & 162    &  165 & 164   \\
2    & 138     & \textbf{149}           & 120     & 117       & 115        & 113          & 109      & 113         & 108  & 130 &  124    \\
3    & 107     & \textbf{114 }          & 66      & 71        & 68         & 77           & 58       & 61          & 59    & 82 &  78    \\
4    & \textbf{58}      & 52            & 24      & 23        & 21         & 18           & 17       & 13          & 11 &  27 & 29 \end{tblr}
}
% \vspace{-1em}
% \vskip -0.1in
\end{table*}
\begin{table}[t!]
\centering
% \caption{\textbf{Comparisons on other benchmarks.} }
\caption{
Comparison with existing VLMs on other benchmarks including VQAv2, GQA, POPE, VisWiz, and SQA. 
Our \textbf{iGVLM} consistently achieves higher accuracy across all backbones, confirming its strong generalization beyond MMStar and MM4.
}
\label{tab:qa}
% \vspace{-1em}
\resizebox{0.95\linewidth}{!}{
\begin{tabular}{l l | c c c c c}
\toprule
% Method & LLM & Res. & PT & IT & VQA-v2 & GQA & VisWiz & SciQA-IMG & TextVQA \\
{Method} & {LLM} & VQAv2 & GQA & POPE & VisWiz  & SQA \\
 % & &  &  &  &  \\
\midrule
{LLaVA-1.5} & Vicuna-7B & 78.5 & 62.0 & 85.4 & 50.0 & 66.8\\

QA-ViT & Vicuna-7B & 79.0 & 62.2 & 85.7 & 51.3 & 68.1\\

DyFo & Vicuna-7B & - & - & 84.8 & -  & 65.5\\

\textbf{\name} & Vicuna-7B & \textbf{79.1} & \textbf{62.8} & \textbf{85.9} & \textbf{52.5} & \textbf{69.9}\\
\midrule
{LLaVA-1.5} & Vicuna-13B & 80.0 & 63.3 & 85.4 & 53.6 & 71.6\\
QA-ViT & Vicuna-13B & 79.9 & 63.2 & 85.8 & \textbf{56.0} & 71.3\\

DyFo & Vicuna-13B & - & - & 84.6 & -& 71.6\\

\textbf{\name} & Vicuna-13B & \textbf{80.2} & \textbf{63.3} & \textbf{86.1} & \underline{55.3} & \underline{71.5}\\
% \textbf{\name-dul} & Vicuna-v1.5-13B & & & & & & & & & \\
\midrule
LLaVA-1.5 & Qwen2.5-3B & 77.7	&62.4	&84.5	&50.7	&72.2 \\
\textbf{iGVLM} &  Qwen2.5-3B & 77.2	& 61.7	&\textbf{84.6}	&\textbf{53.4}	&\textbf{73.0} \\
\bottomrule
\end{tabular}
}
% \vspace{-1em}
% \vskip -0.1in
\end{table}

\subsection{Ablation Study}
\paragraph{Effect of Architectural Components.}
We first examine the contribution of the key design components in \textbf{iGVLM} by ablating (i) instruction-conditioned adaptive layer normalization (AdaLN), (ii) the Zero-FFN adapter used for feature fusion, and (iii) the static branch in the dual-branch architecture.
As shown in Table~\ref{tab:param}, removing AdaLN (\textbf{w/o AdaLN}) leads to consistent performance drops on both MMStar and MM4, indicating that layer-wise, instruction-conditioned normalization is critical for effective visual modulation.
Eliminating the Zero-FFN adapter (\textbf{w/o FFN}) further degrades performance, suggesting that controlled, gradual integration of dynamic features is necessary to avoid disrupting pre-trained visual representations.
The most significant degradation is observed when the static branch is removed (\textbf{w/o Pure}), highlighting the importance of preserving task-agnostic visual priors alongside instruction-guided adaptation.
Together, these results support the hypothesis that instruction-aware perception benefits from explicitly decoupling representation preservation from task-specific modulation.

\begin{table}[t!]
\centering
% \caption{\textbf{{Effect of parameter count.}}}
\caption{
Component ablation of \textbf{iGVLM}. 
We analyze the effect of removing key modules including AdaLN, Zero-FFN, and the static branch.
Performance consistently drops, confirming that instruction-guided modulation, Zero-FFN fusion, and dual-branch structure are all critical for iGVLM’s effectiveness.
}
\label{tab:param}
% \vspace{-1em}
\resizebox{0.95\linewidth}{!}{
\begin{tblr}{
  column{3} = {c},
  column{4} = {c},
  column{5} = {c},
  column{6} = {c},
 hline{2,3} = {-}{},
  hline{1,7}={-}{1pt},
}
Method       & Params(B) & MM4 & MMStar  & VQAv2 & Viswiz \\
LLaVA1.5-13B  & 13.35    & 19  & 32.8      & 80.0  & 53.6   \\
iGVLM-13B     & 13.78    & \textbf{23}  & \textbf{36.4}      & \textbf{80.2}  & \textbf{55.3}\\
-w/o AdaLN & 13.78   & 22  & 35.1       & 80.2  & 53.5   \\
-w/o FFN & 13.78 & 17 & 34.1 & 80.1 & 54.7\\
-w/o Pure & 13.48 & 5 & 27.3 & 60.2 & 37.9\\
\end{tblr}
}

% \vspace{-1em}
% \vskip -0.1in
\end{table}

\begin{table}[t!]
\centering
% \caption{\textbf{{Different variants of iGVLM.}} }
\caption{
Comparison of \textbf{iGVLM} variants on multiple benchmarks. 
\textbf{iGVLM-MoF} and \textbf{iGVLM-Cross} replace AdaLN with interleaved or cross-attention modulation, 
but both yield lower accuracy, highlighting the efficiency of our AdaLN-based design.
}
\label{tab:variant}
% \vspace{-1em}
\resizebox{0.95\linewidth}{!}{
\begin{tabular}{l  |ccccc}
\toprule
Method &  MMStar & VQAv2& GQA &VisWiz  & POPE \\
\midrule
iGVLM&{34.7} & 79.1& 62.8& 69.9 &{85.9}\\
\midrule
iGVLM-MoF &	32.0 & 79.3	&62.7 &	68.8		&85.4	 \\
iGVLM-Cross & 33.0 & 79.4	&62.8&	68.6&	85.8\\
\bottomrule
\end{tabular}
}
% \vspace{-1em}
% \vskip -0.1in
\end{table}

\paragraph{Comparison with Alternative Modulation Designs.}
We further compare iGVLM with two representative variants that adopt different strategies for integrating instruction signals into visual features.
\textbf{iGVLM-MoF}~\cite{tong2024eyes} interleaves static and dynamic tokens, while \textbf{iGVLM-Cross}~\cite{peebles2023scalable} replaces AdaLN with cross-attention-based interaction.
As reported in Table~\ref{tab:variant}, both variants underperform the original iGVLM.
MoF weakens the explicit separation between static and dynamic representations, while cross-attention introduces additional computational overhead and optimization noise without improving instruction consistency.
These comparisons suggest that AdaLN-based modulation offers a more effective and efficient mechanism for conditioning visual representations on textual instructions.

\paragraph{Scaling Behavior.}
Finally, we analyze how model capacity influences instruction-aware reasoning by training \textbf{iGVLM-1.5B} based on the Qwen2.5-1.5B backbone and comparing it with larger variants.
As shown in Table~\ref{tab:scaling}, even the smallest iGVLM model improves over its LLaVA-1.5B counterpart on MMStar (19.7 vs.~17.1), indicating that instruction-guided visual modulation is beneficial across model scales.
However, the lower MM4 score of iGVLM-1.5B (16 vs.~19) reveals that consistent multi-instruction reasoning requires sufficient language modeling capacity.
Performance improves monotonically from 1.5B to 3B, 7B, and 13B, and notably, iGVLM-3B outperforms the larger Vicuna-13B variant on MM4.
This trend suggests a strong synergy between the proposed dual-branch vision encoder and modern language backbones, and highlights that instruction-aware visual reasoning is jointly constrained by visual modulation and language capacity.

% \paragraph{Scaling Behavior.}
% To analyze the effect of model capacity, we further train a lightweight variant, \textbf{iGVLM-1.5B}, based on the Qwen2.5-1.5B backbone, and compare it with larger configurations under both MMStar and MM4 benchmarks. 
% As summarized in~\Cref{tab:scaling}, iGVLM-1.5B achieves a noticeable improvement over LLaVA-1.5B on MMStar (19.7 vs.~17.1), demonstrating that even under limited capacity, instruction-guided modulation effectively enhances single-instruction understanding. 
% However, its MM4 score (16 vs.~19) falls slightly behind, suggesting that consistent reasoning across multiple queries requires sufficient contextual modeling capacity within the language backbone.

% As the model scales from 1.5B to 3B, 7B, and 13B, we observe a clear monotonic increase in both MMStar and MM4 performance. 
% Notably, \textbf{iGVLM-3B} already surpasses iGVLM-13B and other larger Vicuna-based variants on MM4 (29 vs.~23), highlighting the strong synergy between our dual-branch vision encoder and the more advanced Qwen2.5 architecture. 
% This trend indicates that the effectiveness of instruction-guided visual modulation is not limited to model size alone but also depends on the expressiveness of the underlying LLM family.

% Overall, these results confirm that iGVLM scales gracefully across capacities and architectures, maintaining competitive efficiency while leveraging stronger language backbones to achieve more coherent multi-instruction reasoning.

\begin{table}[t]
\centering
\caption{Scaling performance of iGVLM on MMStar and MM4 benchmarks under different model sizes and LLM backbones.}
\label{tab:scaling}
% \vspace{-1em}
\scalebox{0.85}{
\begin{tabular}{l l c c}
\toprule
\textbf{Model} & \textbf{LLM} & \textbf{MMStar (Avg.)} & \textbf{MM4 (Score)} \\
\midrule
LLaVA-1.5B   & Qwen2.5-1.5B  & 17.1 & 19 \\
iGVLM-1.5B   & Qwen2.5-1.5B  & 19.7 & 16 \\
iGVLM-3B     & Qwen2.5-3B    & 21.3 & 29 \\
iGVLM-7B     & Vicuna-7B     & 34.7 & 17 \\
iGVLM-13B    & Vicuna-13B    & 36.4 & 23 \\
\bottomrule
\end{tabular}
}
\vspace{-1em}
% \vskip -0.1in
\end{table}

% \section{Conclusion}
% In this paper, we introduce a novel vision-language model named iGVLM, which leverages input instructions to guide the generation of visual features and enhance the image understanding capabilities of large-scale multimodal models. In a series of multimodal benchmark tasks, iGVLM consistently demonstrates robust performance. Our approach yields significant improvements across various model scales, showcasing its scalability. To facilitate a more comprehensive evaluation of multimodal models, we also introduce a new test set, MM4, which assesses models' ability to understand images from multiple perspectives and fills a critical gap in the field.

\section{Conclusion}

We presented \textbf{iGVLM}, a decoupled instruction-guided vision encoder that enables visual representations to be modulated according to textual instructions without retraining the visual backbone.
By explicitly separating representation preservation from instruction-conditioned adaptation, iGVLM provides an efficient and stable mechanism for question-aware visual perception in vision--language models.
Extensive experiments across diverse benchmarks demonstrate that iGVLM consistently improves instruction sensitivity and fine-grained multimodal reasoning while maintaining strong general-purpose performance across model scales from 3B to 13B parameters.
In addition, we introduced \textbf{MM4}, a controlled diagnostic benchmark for evaluating multi-instruction, multi-query visual reasoning, enabling targeted analysis of instruction-conditioned perception.
Overall, our results highlight the importance of explicitly conditioning the utilization of visual features on linguistic instructions, and suggest decoupled visual modulation as a principled design direction for instruction-aware multimodal models.

\nocite{langley00}

\bibliography{example_paper}
\bibliographystyle{icml2026}

%%%%%%%%%%%%%%%%%%%%%%%%%%%%%%%%%%%%%%%%%%%%%%%%%%%%%%%%%%%%%%%%%%%%%%%%%%%%%%%
%%%%%%%%%%%%%%%%%%%%%%%%%%%%%%%%%%%%%%%%%%%%%%%%%%%%%%%%%%%%%%%%%%%%%%%%%%%%%%%
% APPENDIX
%%%%%%%%%%%%%%%%%%%%%%%%%%%%%%%%%%%%%%%%%%%%%%%%%%%%%%%%%%%%%%%%%%%%%%%%%%%%%%%
%%%%%%%%%%%%%%%%%%%%%%%%%%%%%%%%%%%%%%%%%%%%%%%%%%%%%%%%%%%%%%%%%%%%%%%%%%%%%%%
\newpage
\appendix
\onecolumn

% \section{Supplementary Material}
\section{Hyperparameters}
Our iGVLM follows the hyperparameter configuration of LLaVA-1.5. Since iGVLM employs a dual-branch vision encoder, we set the learning rate to 6e-4 during the pre-training stage. The hyperparameters for the first-stage vision–language alignment pre-training and the second-stage instruction tuning are provided in~\cref{tab:settings}.
\begin{table}[t]
    \centering
    \caption{Hyperparameters of iGVLM are the same as the
original LLaVA-1.5, except the learning rate in pretraining.}
\label{tab:settings}
   \begin{tabular}{l|cc} 
\hline
Hyperparameter  & \multicolumn{1}{l}{Pretrain} & \multicolumn{1}{l}{Finetune}  \\ 
\hline
Batch size      & 256                          & 128                           \\
Lr             & 6e-4                        & 2e-5                          \\
Lr schedule     & \multicolumn{2}{c}{cosine decay}                             \\
Lr warmup ratio & \multicolumn{2}{c}{0.03}                                     \\
Weight decay    & \multicolumn{2}{c}{0}                                        \\
Epoch           & \multicolumn{2}{c}{1}                                        \\
Optimizer       & \multicolumn{2}{c}{AdamW}                                    \\
DeepSpeed stage & 2                            & 3                             \\
\hline
\end{tabular}
\vskip -0.1in
\end{table}

\section{Analyzing Question Diversity via CLIP Text Embeddings}
\label{text_visualization}
\begin{figure}[t!]
\vskip 0.2in
    \centering
    % 第一行
    \includegraphics[width=0.3\linewidth]{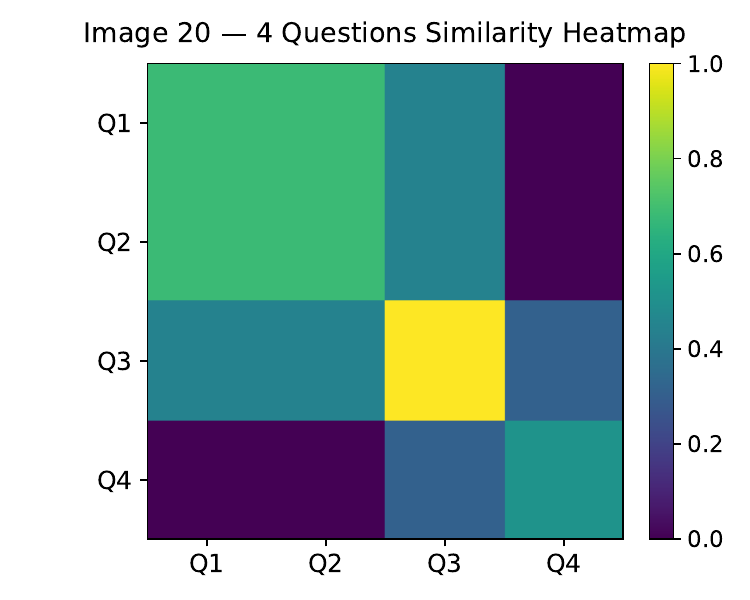}
    %\hspace{0.5mm}%
    \includegraphics[width=0.3\linewidth]{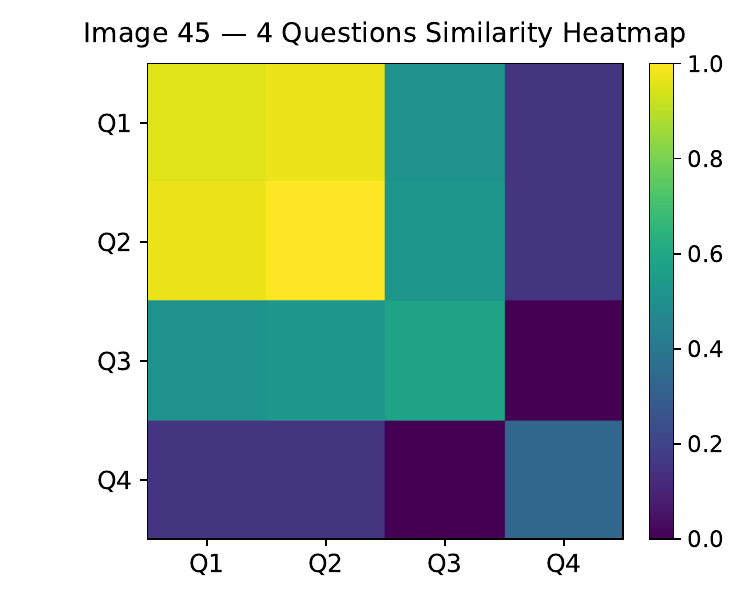} %\hspace{0.5mm}%
 \includegraphics[width=0.3\linewidth]{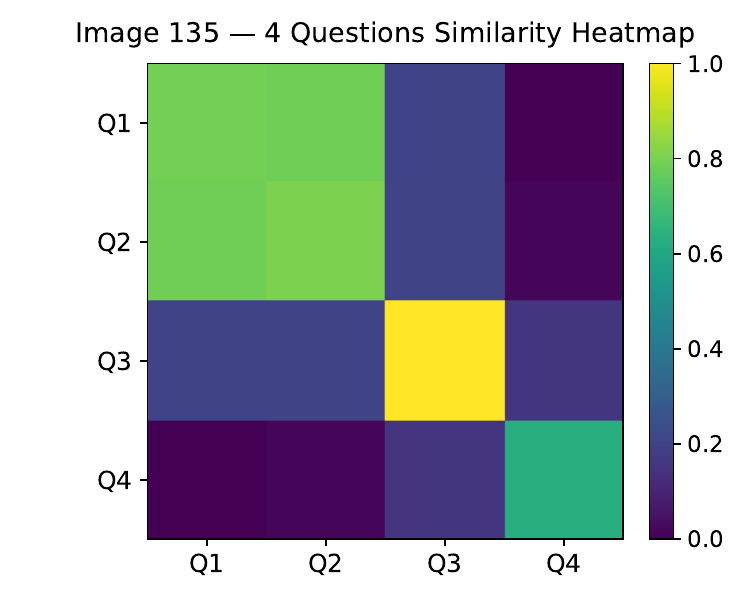}
    \\[-0.5mm]
    % 第二行
    \includegraphics[width=0.3\linewidth]{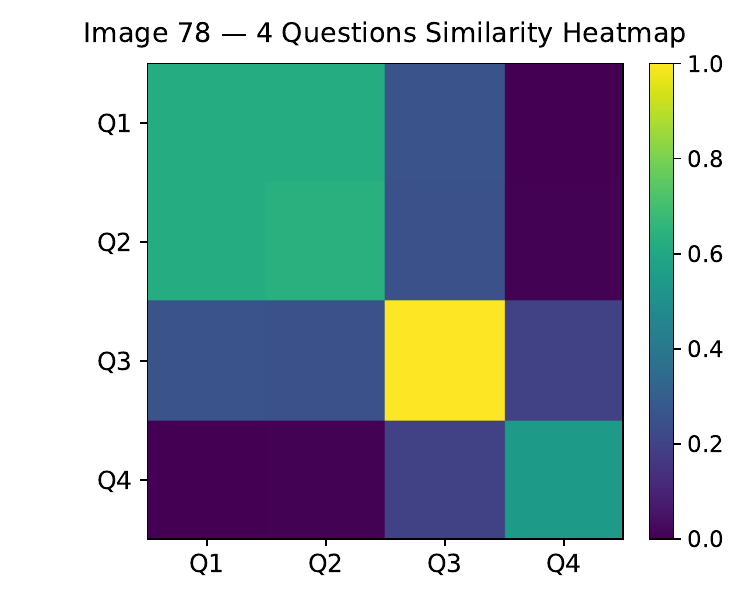}\hspace{0.5mm}%
    \includegraphics[width=0.3\linewidth]{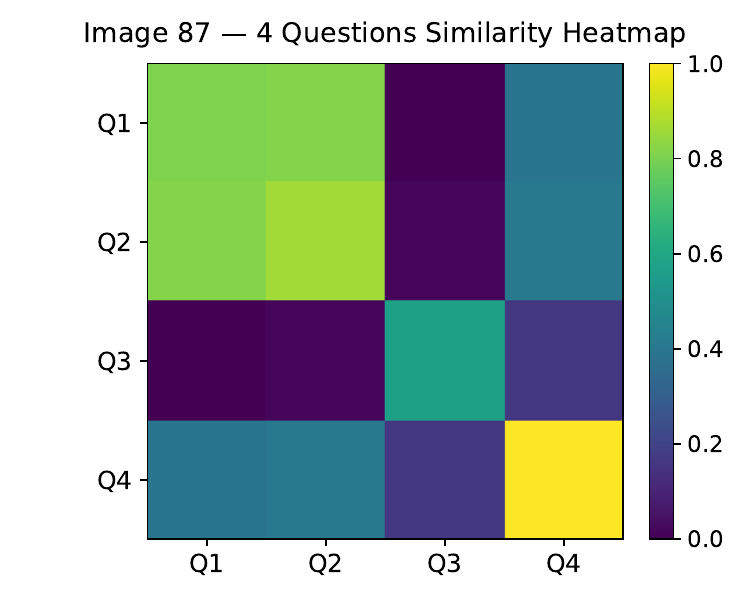}\hspace{0.5mm}%
    \includegraphics[width=0.3\linewidth]{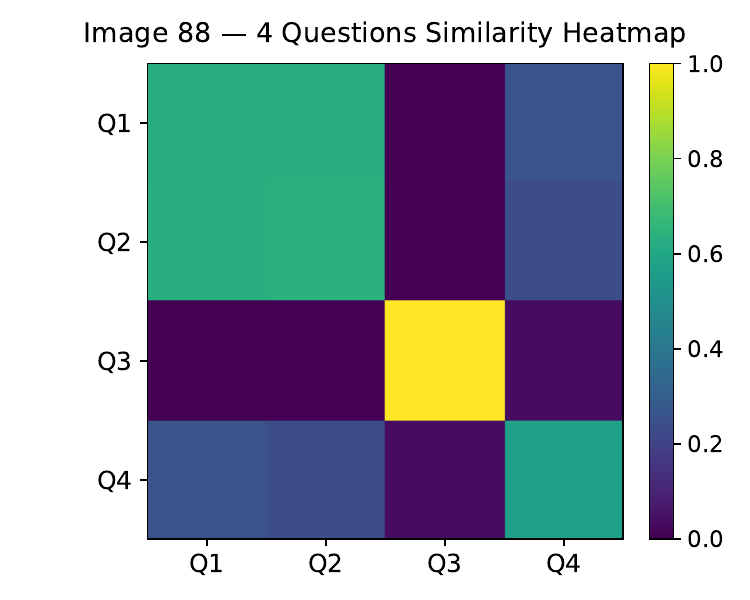}
    \\[-0.5mm]
    % 第三行
    \includegraphics[width=0.3\linewidth]{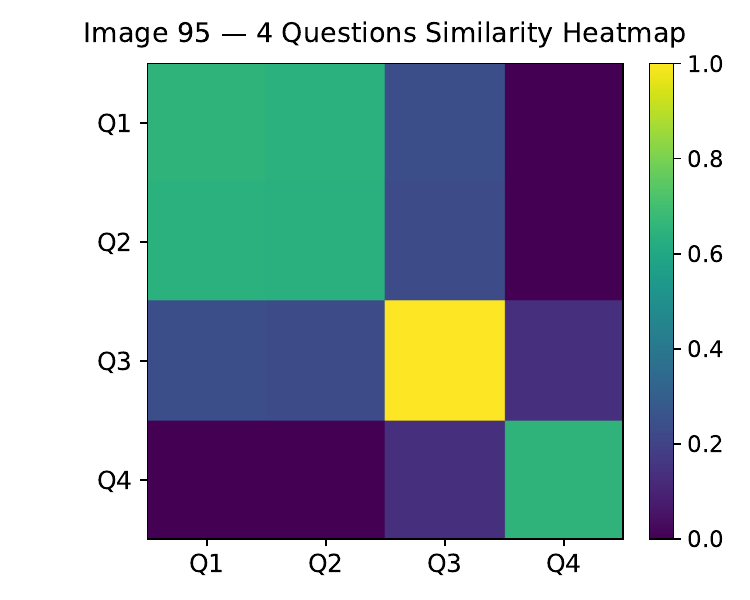}\hspace{0.5mm}%
    \includegraphics[width=0.3\linewidth]{visual/image_135_similarity.pdf}\hspace{0.5mm}%
    \includegraphics[width=0.3\linewidth]{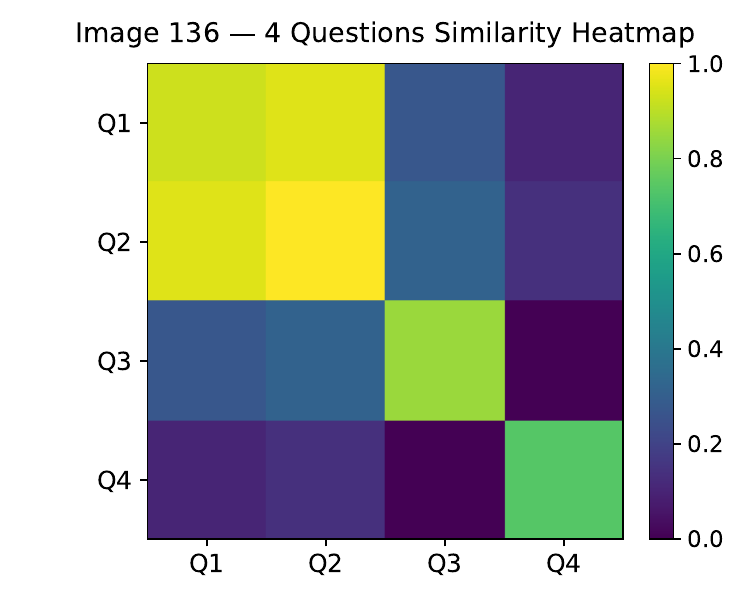}

    \caption{Different questions similarity heatmap.}
    
    \label{fig::visual}
\end{figure}

During the construction of the MM4 benchmark, we enforce strict criteria to ensure sufficient distinction between question pairs $(Q_1, Q_2)$ and $(Q_3, Q_4)$. 
To further verify the semantic separation achieved by these rules, we extract feature representations for all questions using the CLIP text encoder, compute pairwise cosine similarities, and visualize the results. 
We randomly select nine representative images for illustration. 
As shown in~\cref{fig::visual}, each heatmap consistently exhibits three distinct color clusters along both rows and columns, indicating that our rule-based construction effectively maintains semantic diversity and minimizes redundancy among questions.

\section{More Cases of MM4}
\label{sec:cases}
We present additional examples from MM4 to provide a more intuitive demonstration of the importance of uestion-aware understanding for vision-language models.

\begin{figure*}[t]\centering
\includegraphics[width=0.98\linewidth]{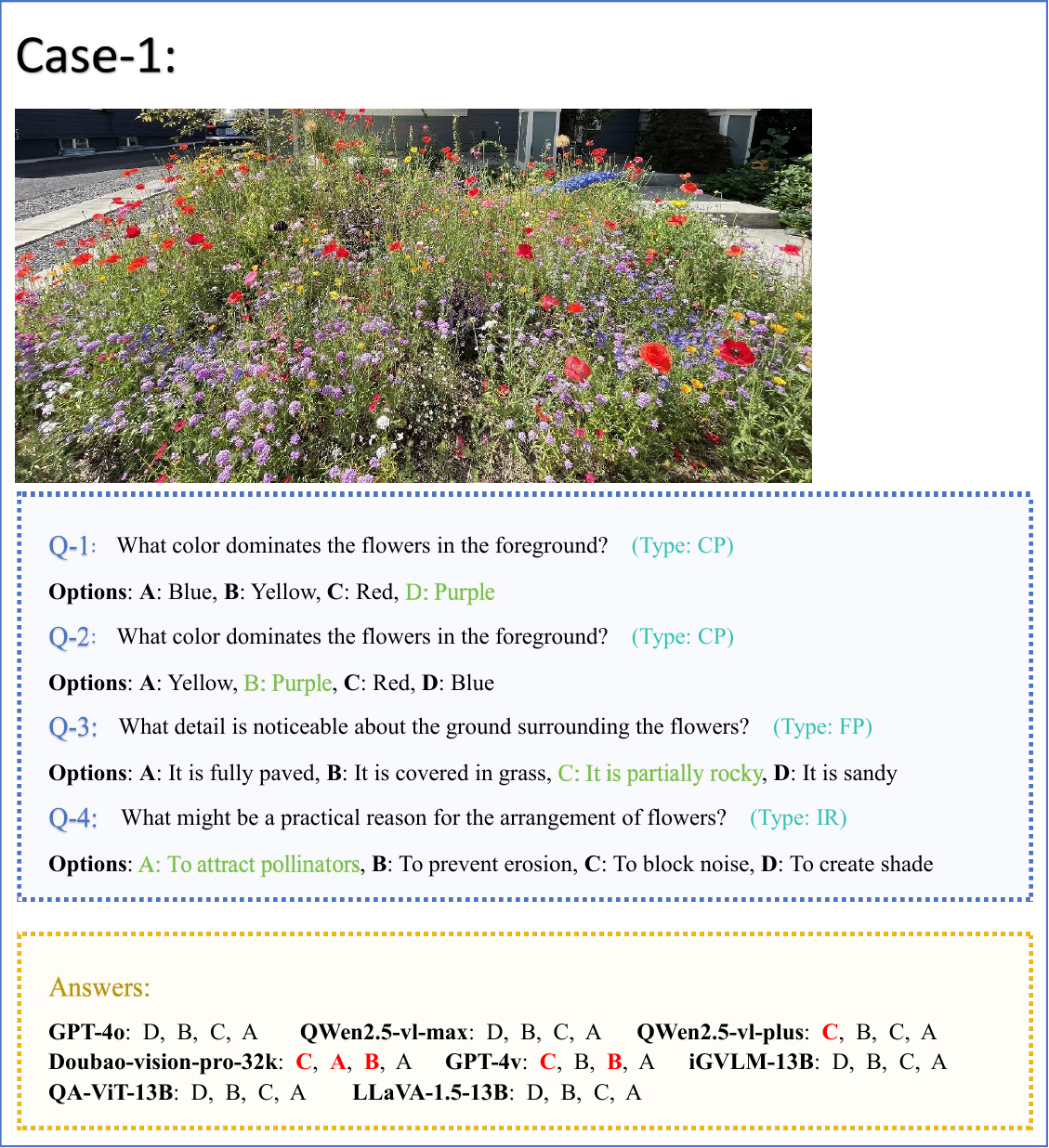}
\label{case1}
\end{figure*}

\begin{figure*}[t]\centering
\includegraphics[width=0.98\linewidth]{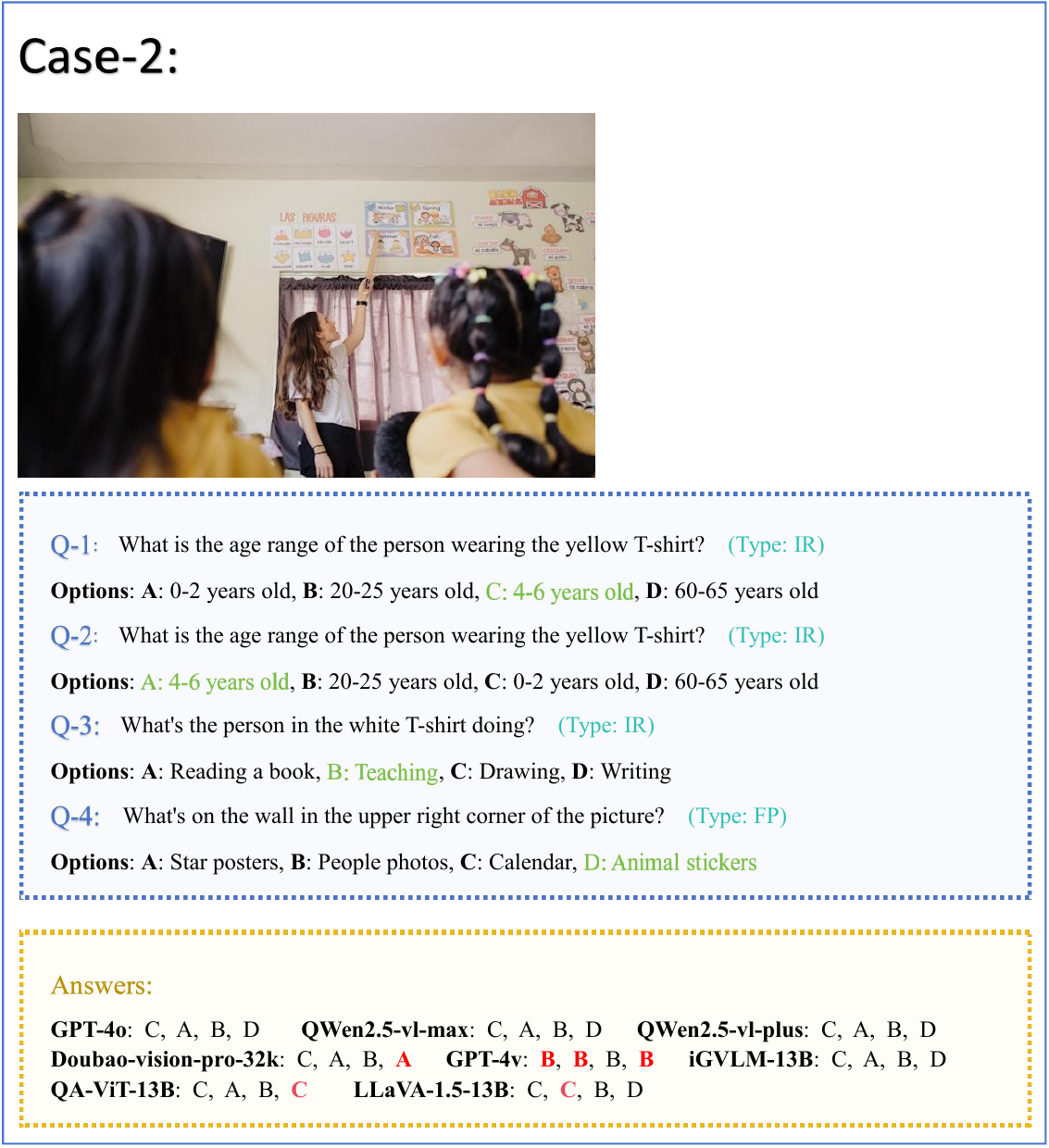}
\label{case2}
\end{figure*}

\begin{figure*}[t]\centering
\includegraphics[width=0.98\linewidth]{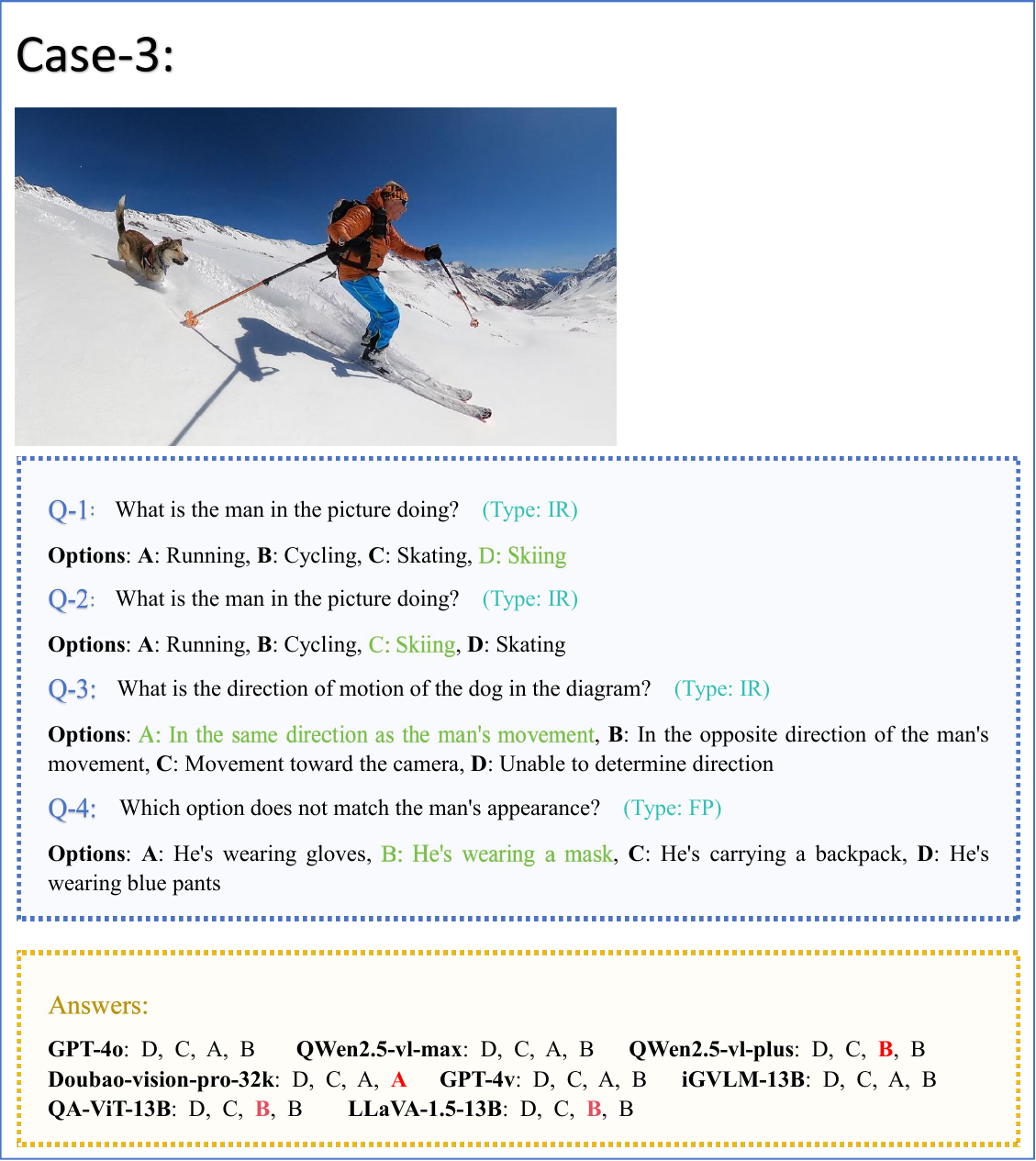}
\label{case3}
\end{figure*}

\begin{figure*}[t]\centering
\includegraphics[width=0.98\linewidth]{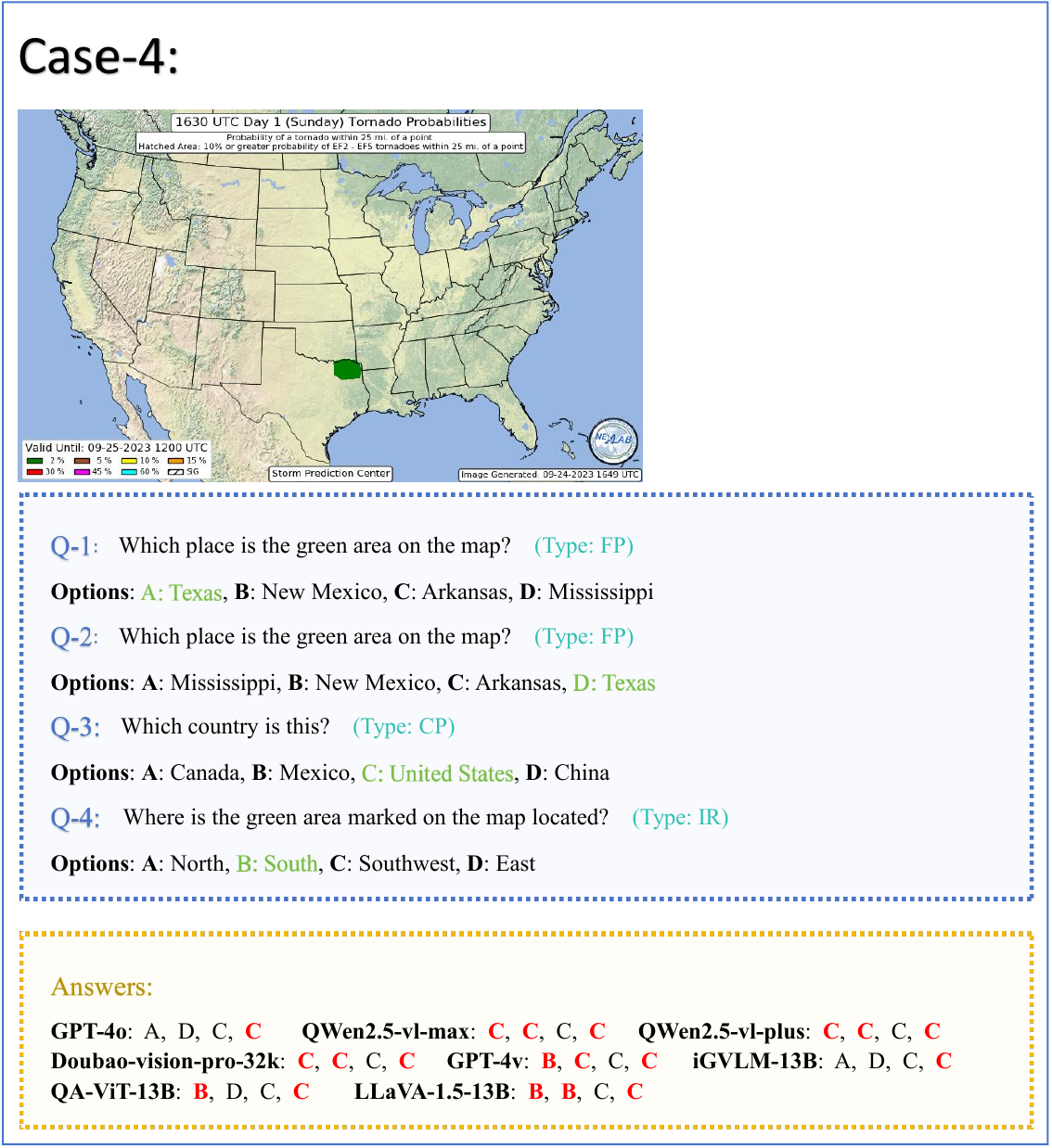}
\label{case4}
\end{figure*}

\begin{figure*}[t]\centering
\includegraphics[width=0.98\linewidth]{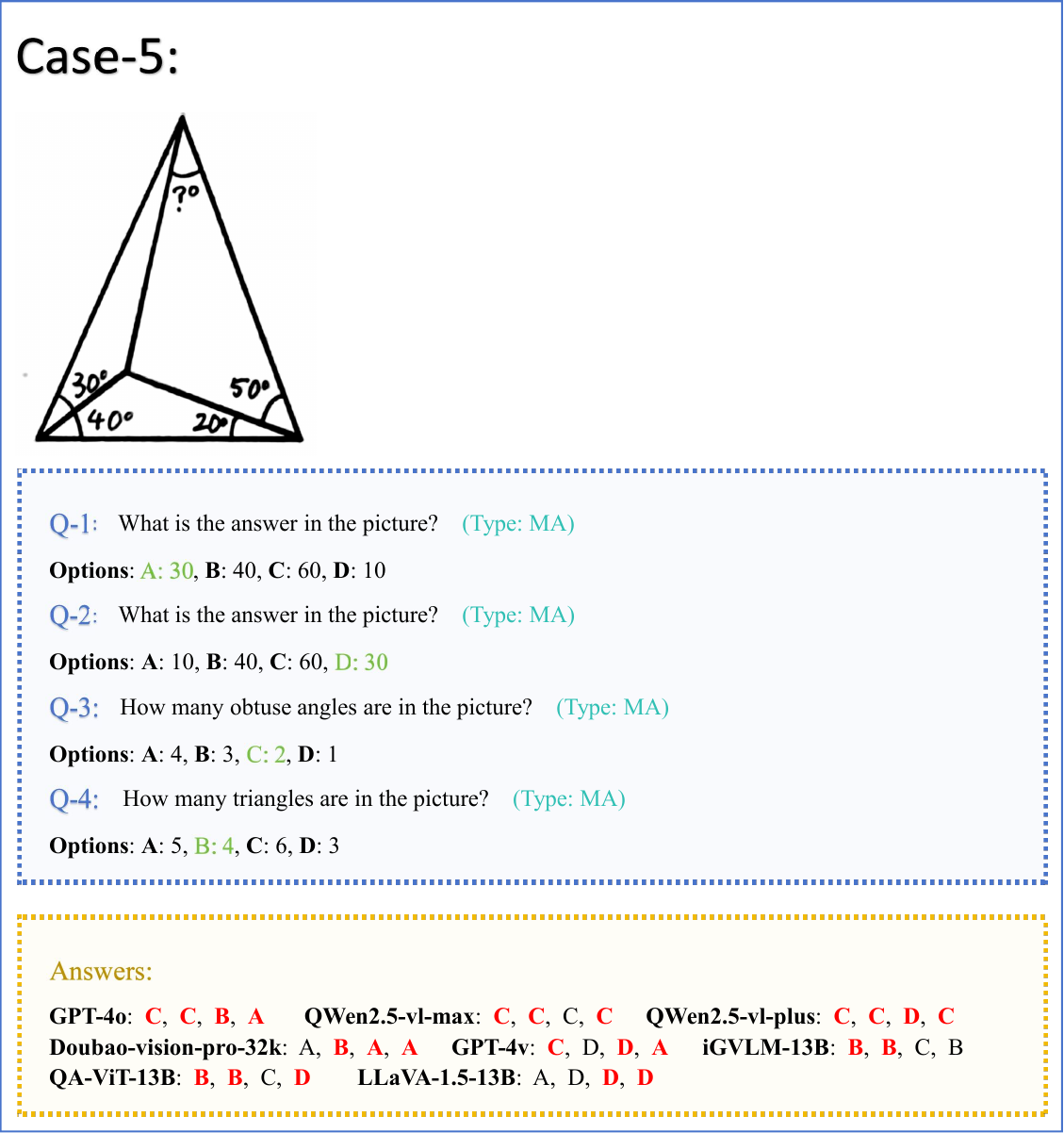}
\label{case5}
\end{figure*}

\begin{figure*}[t]\centering
\includegraphics[width=0.98\linewidth]{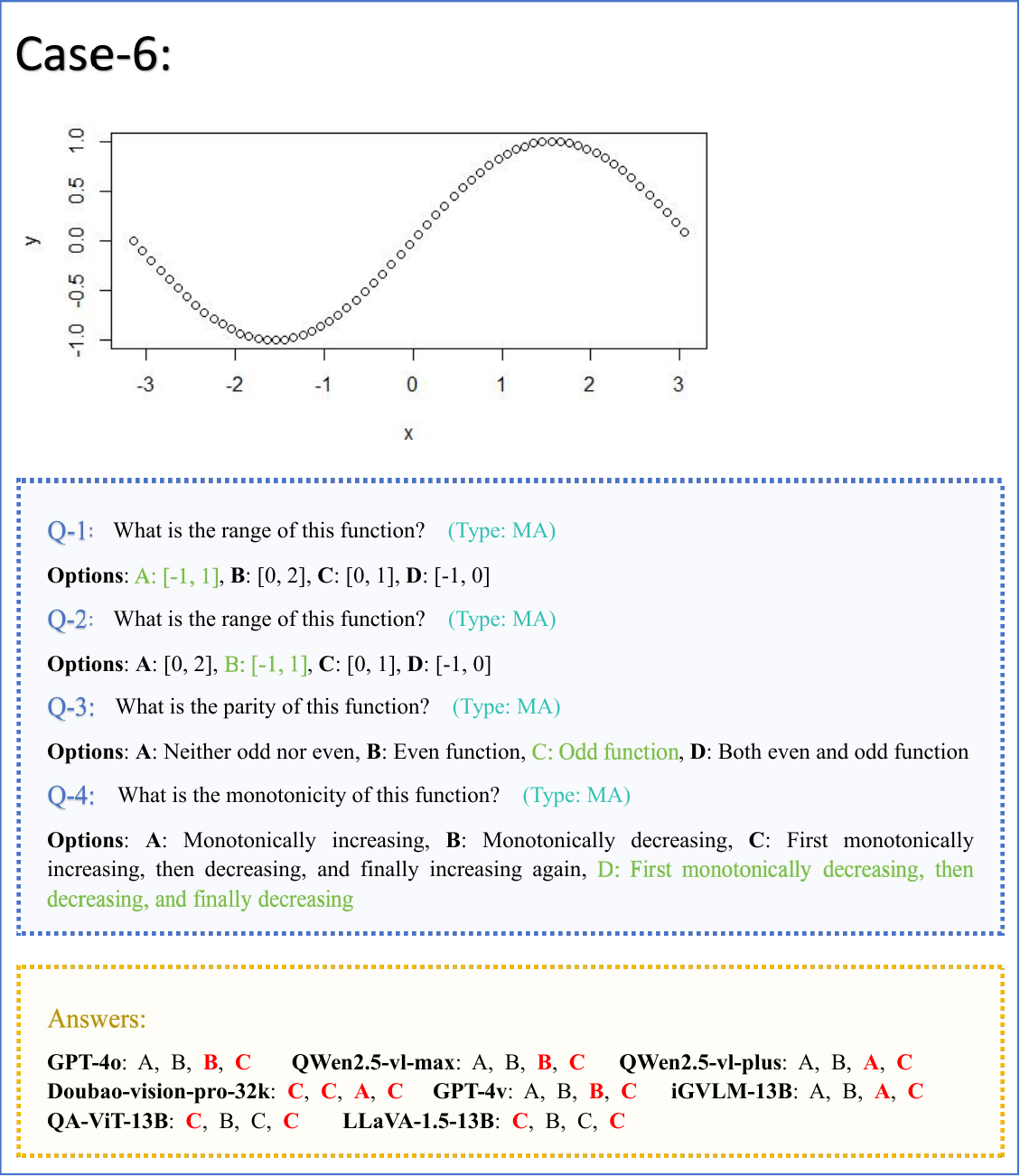}
\label{case6}
\end{figure*}

%%%%%%%%%%%%%%%%%%%%%%%%%%%%%%%%%%%%%%%%%%%%%%%%%%%%%%%%%%%%%%%%%%%%%%%%%%%%%%%
%%%%%%%%%%%%%%%%%%%%%%%%%%%%%%%%%%%%%%%%%%%%%%%%%%%%%%%%%%%%%%%%%%%%%%%%%%%%%%%

\end{document}